\documentclass{article}

\usepackage[preprint]{neurips_2025}

\usepackage[utf8]{inputenc} 
\usepackage[T1]{fontenc}    
\usepackage{hyperref}       
\usepackage{url}            
\usepackage{booktabs}       
\usepackage{amsfonts}       
\usepackage{nicefrac}       
\usepackage{microtype}      
\usepackage{xcolor}         
\usepackage{algorithm}
\usepackage{algpseudocode}
\usepackage{graphicx}
\usepackage{amsmath}
\usepackage{float}
\usepackage{wrapfig}

\usepackage{xcolor}
\pagecolor{white}

\title{Behind the Noise: Conformal Quantile Regression Reveals Emergent Representations}

\author{%
  Petrus H. Zwart\thanks{[Corresponding Author] Center for Advanced Mathematics in Energy Research Applications; Berkeley Synchrotron Infrared Structural Biology program; Molecular Biophysics and Integrated Bioimaging Division, Lawrence Berkeley National Laboratory} \\
  Berkeley, CA 94720, USA \\
  \texttt{PHZwart@lbl.gov} \\
  \And
  Tamas Varga\thanks{Environmental Molecular Science Laboratory, Pacific Northwest National Laboratory} \\
  Richland, WA 99354, USA \\
  \texttt{Tamas.Varga@pnnl.gov} \\
  \And
  Odeta Qafoku\footnotemark[2] \\
  Richland, WA 99354, USA \\
  \texttt{Odeta.Qafoku@pnnl.gov} \\
  \AND
  James A. Sethian\thanks{Center for Advanced Mathematics in Energy Research Applications; Applied Mathematics and Computational Research Division, Lawrence Berkeley National Laboratory; Department of Mathematics, University of California, Berkeley} \\
  Berkeley, CA 94720, USA \\
  \texttt{JASethian@lbl.gov} \\
}

\begin{document}

\maketitle

\begin{abstract}
Scientific imaging often involves long acquisition times to obtain high-quality data, especially when probing complex, heterogeneous systems. However, reducing acquisition time to increase throughput inevitably introduces significant noise into the measurements. We present a machine learning approach that not only denoises low-quality measurements with calibrated uncertainty bounds, but also reveals emergent structure in the latent space. By using ensembles of lightweight, randomly structured neural networks trained via conformal quantile regression, our method performs reliable denoising while uncovering interpretable spatial and chemical features -- without requiring labels or segmentation. Unlike conventional approaches focused solely on image restoration, our framework leverages the denoising process itself to drive the emergence of meaningful representations. We validate the approach on real-world geobiochemical imaging data, showing how it supports confident interpretation and guides experimental design under resource constraints.
\end{abstract}

\section{Introduction}
In geobiochemical sciences, X-ray Computed Tomography (XCT) and Scanning Electron Microscopy with Energy Dispersive X-ray Spectroscopy (SEM-EDX) are widely used imaging modalities for characterization of complex, heterogeneous samples, such plants, roots, as soil. XCT provides three-dimensional structural information at the micrometer scale, revealing features like porosity and water permeability. In contrast, SEM-EDX offers high-resolution chemical data through elemental distribution maps of cross-sectional areas of the sample. When combined with additional data, such as information on the soil microbiome, these imaging techniques form the foundation for building a digital model of soil. This model is intended to offer mechanistic insights into fundamental geobiological processes that are crucial for understanding agriculture, ecology, and climate change. Imaging samples using SEM-EDX and XCT is a time-intensive endeavor. Where a typical XCT imaging of a soil core takes around 6 hours on a lab-based instrument, a typical SEM-EDX runs takes over 5 hours per $mm^2$ to yield satisfactory signal to noise ratios. 

In both cases, the time required to obtain sufficient data for downstream scientific analysis is significant. Sample deterioration, or simply not having enough access to the required instrumentation makes certain experiments practically impossible. In the case of SEM-EDX for instance, imaging a square centimeter of soil can take over 48 hours, but will only yield an image with a suboptimal signal to noise ratio.  In the case of XCT, gravity induced local settling of the samples can introduce artifacts that limit the final results. To address these challenges, we investigate if the use of fast, high-noise measurements with a subsequent denoising and uncertainty quantization step can provide actionable insights to design new, high-value experiments, overcoming limitations imposed by laboratory resource constraints. 

While traditional denoising methods produce only point estimates, this is often insufficient in scientific imaging, where results must be accompanied by calibrated uncertainty estimates to assess reliability and avoid overinterpretation. In supervised settings, denoising is further complicated by limited training data, making it difficult to build generalizable models for rare or complex samples. We present a lightweight, ensemble-based denoising framework that incorporates built-in uncertainty quantification. By assembling random neural networks with low parameter counts, we leverage variability across models to produce robust predictions. Rather than using ensemble spread to approximate uncertainty, we directly train each network to predict quantiles using independent projection heads connected to a shared convolutional backbone. This shared latent space promotes feature reuse, avoids quantile crossing, and removes the need for separate estimators. Beyond denoising, we show that the latent representations learned by the ensemble self-organize into meaningful morphological and chemical spatial features without providing explicit segmentation labels at training time.

We demonstrate the effectiveness of this approach on simulated data to gauge the effect of hyper parameter choices on the denoising setup, and on experimental XCT and SEM-EDX data to probe its performance in a controlled real-world scenario.

\section{Background}
Neural networks have demonstrated strong performance in scientific imaging tasks, especially for learning and operating on complex spatial patterns. However, their effectiveness often hinges on access to large training datasets, which can be infeasible in many scientific domains \citep{litjens2017survey, albert2020deep, gupta2022deep}. When data is scarce, overfitting and poor generalization are common. To address this, regularization strategies and ensembling methods can be deployed.

Ensembles are a robust approach to reducing predictive variance and improving model reliability. By training multiple models and averaging their predictions, ensemble methods average out errors, providing more stable outputs. Techniques range from simple averaging to more structured approaches such as boosting \citep{freund1997decision, bentejac2021comparative} and Stochastic Weight Averaging \citep{yang2019swalp, guo2023stochastic}. 

In addition to robustness, ensembles can offer a route to uncertainty quantification (UQ). However, the spread in ensemble predictions typically reflects confidence in a point estimate (e.g., an L$_2$ minimum), and may not fully capture the shape or tails of the underlying predictive distribution \citep{lakshminarayanan2017simple}. Alternatives such as Gaussian Process Regression (GPR) or Quantile Regression (QR) provide more direct control over distributional uncertainty. GPR offers flexible, Bayesian modeling but scales poorly and typically assumes properties \-- such as stationarity \--  that might be hard to satisfy in a real-world scenarios \citep{Rasmussen2005-fw, Noack2023}. QR, used here, avoids these assumptions and estimates predictive intervals instead of a full distribution by minimizing the pinball loss \citep{Steinwart2011-rk}.

To calibrate the uncertainty estimates produced by quantile regression, we apply \emph{conformal prediction} (CP), a distribution-free framework that guarantees coverage of prediction intervals under minimal assumptions \-- the exchangeability of data points \citep{shafer2008tutorial,Fontana2023-uc,Papadopoulos2024-if}. CP adjusts the predicted quantile bounds based on held-out calibration data, ensuring that the resulting intervals have statistically valid coverage \citep{Sesia2020-as}. This allows one to move from approximate, model-dependent uncertainty estimates to confidence intervals that can be trusted by downstream applications, regardless of model architecture or training dynamics.

While the main goal of our approach is to produce reliable, uncertainty-aware predictions from noisy data, we observe a compelling secondary effect: the internal representations learned by our models exhibit emergent structure. Though trained only to denoise and predict quantiles, the ensemble of networks develop combined latent space that consistently organizes morphologically and chemically distinct features \-- effectively discovering segmentation-like partitions without human-guided supervision.

This phenomenon aligns with findings in recent self-supervised representation learning, where networks trained on pretext tasks like image reconstruction or contrastive prediction also learn features that align with human-interpretable concepts. Methods such as DINO \citep{caron2021emerging}, MAE \citep{he2022masked}, and the Segment Anything Model (SAM) \citep{kirillov2023segment} show that carefully constructed objectives and architectural biases are sufficient for structure to emerge from unlabeled data. Our findings extend this principle: even small, randomly structured networks trained for uncertainty-aware denoising can yield latent spaces that are both informative and semantically meaningful.

\section{Methodology}
Here we will demonstrate that an ensemble of neural networks can achieve robust uncertainty estimation in scientific imaging applications. By leveraging ensembles of parameter-lean convolutional neural networks with a random architecture \cite{Roberts:yr5117} in a quantile regression and conformal prediction approach, we can construct a system that mitigates overfitting, enhances generalization and provides well-calibrated uncertainty estimates. This approach promises reliability in both the accuracy and reliability off predictions, even in data-limited scenarios.

\subsection{Network Design}
In order to perform QR of a number of specified quantiles in an parameter efficient fashion, a network was constructed that consists of three independent projector heads tethered to a multidimensional feature vector generated by a base convolutional neural network (CNN) acting upon an image. This network design allows one to reuse generated image features for a number of different quantiles, reducing the computational burden, and likely enforce an encoding that is sufficient general to carry semantically distinct interpretations. The projector heads for the median directly estimates the median, while the projection heads for the lower and upper quantiles are positive offset parameters, to be subtracted off or added to the median to get the lower and upper quantiles. This strategy avoids quantile crossing, while domain restrictions are enforced via clamping operators. Instead of linearly combining the different pinball losses for each quantile we use batch-wise stochastic task switching, optimizing either the lower quantile, upper quantile or median for each batch. Because the base network is connected to each projector head, the weights and biases for the CNN are optimized during each batch, while the parameters for individual projector heads are updated at random.

\subsection{Stochastic Network Generation}

A well‐known challenge in building neural‐network ensembles is making sure the individual models are different enough to bring complementary strengths. In CNNs, diversity usually comes from random weight initialization, data‐augmentation strategies, or tweaking hyperparameters. Critically though, the core network architecture stays the same. Here, we take a different tack: we assemble an ensemble whose members have genuinely different wiring diagrams. Inspired by Mixed Scale Dense Networks (MSDNet) \cite{pelt2018mixed,pelt2018improving}, we define each network by a randomly generated single‐source, single‐sink directed acyclic graph (SS‐DAG). Nodes in this graph collect and combine feature maps; edges carry out convolutions (with dilations picked at random) and activations. By varying the number of nodes, our \emph{depth} hyperparameter, and the rules that govern how many edges each node emits and how long those edges are, we can flexibly trade off complexity, parameter count, and effective receptive field.
The networks are constructed according to a procedure outline in the supplementary materials, where the underlying structure of the network graph is governed by two hyperparameters.
First, we let the node degree follow a power‐law, controlled by parameter $\gamma$
\begin{equation}
P(k) \propto k^{-\gamma},
\end{equation}
so that higher $\gamma$ yields sparser connectivity on average. Second, we bias skip‐connection lengths with parameter $\alpha$
\begin{equation}
P\bigl(\mathrm{edge}(i,j)\bigr)\propto \exp\bigl(-\alpha\,|i-j|\bigr),
\end{equation}
\begin{figure}[!h]
\begin{center}
\includegraphics[width=0.995\textwidth]{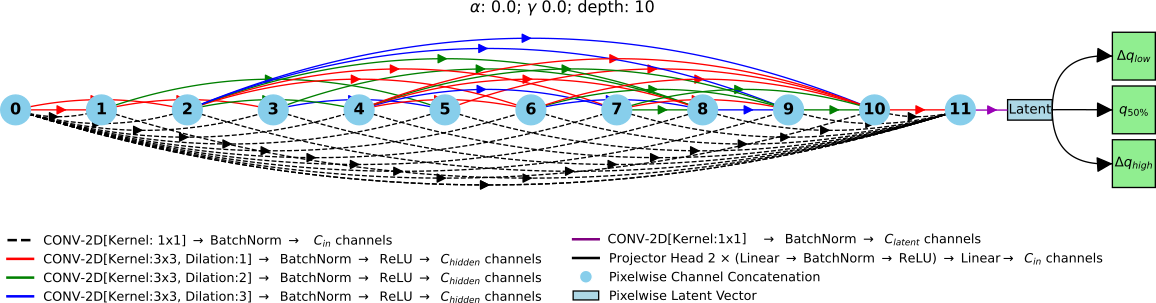}
\end{center}
\caption{An instance of a random sparse mixed-scale network (SMSNet) with tethered quantile projection heads. Arrows represent convolutional kernels and activation functions, nodes are feature channel concatenators. The the bottom (black) skip connections connect the input and hidden layers directly to the output. The top (colored) connections between nodes are formed vi a stochastic process.}
\label{fig:1}
\end{figure}
Larger $\alpha$ discourages long jumps and constructs graph that contain longer, more densely connected paths. Once the SS‐DAG is drawn, nodes are sorted topologically, yielding a execution sequence that guarantees that the required feature maps are available when the image passes through the computational graph. Each \emph{Sparse Mixed Scale Network} (SMSNet) is compact yet structurally unique, Figure \ref{fig:1}, and when pooled into an ensemble, the varied architectural layouts offer complementary views of the input data, boosting robustness and denoising performance beyond what homogeneous ensembles can achieve \cite{Zwart2025}. Note that individual networks do not directly denoise the image, but act as a feature encoder, mapping each pixel and its receptive field into a compact latent vector of user configurable size, e.g.\ \(d=8\)). These latent representations not only support downstream tasks such as quantile estimation but also encode semantically meaningful information about the underlying data. As shown later, clustering these latent vectors reveals coherent groupings that correspond to structural and compositional features, suggesting that the model captures high-level abstractions—such as boundaries, texture, or material type—without requiring explicit supervision.

\subsection{Quantile Projection Heads}
To produce calibrated uncertainty estimates, each SMSNet is followed by a set of \emph{quantile projection heads} that map its latent representation to predicted quantiles of the target distribution.  Let $z \in \mathbb{R}^d$ be the $d$-dimensional latent vector (user‐configurable; e.g.\ $d=8$).  For each quantile level $q\in(0,1)$, we attach an independent multilayer perceptron (MLP) head $h_q(z)$ to the base network.  A typical configuration uses three fully connected layers, reducing the latent dimension to the desired dimension only in the final projection layer \-- this behavior can be altered upon instantiation of the network.

Rather than training each head completely independently (which can lead to quantile crossing \cite{bondell2010noncrossing,koenker2005quantile}), we parameterize the lower‐ and upper‐quantile outputs as nonnegative offsets around the median prediction $\hat y_{0.5}$.  Denote the raw MLP outputs by $\tilde y_{q_{\mathrm{low}}}$ and $\tilde y_{q_{\mathrm{high}}}$.  We then compute
\begin{equation}
\begin{aligned}
\hat y_{q_{\mathrm{low}}} &= \hat y_{0.5} - \mathrm{softplus}\bigl(\tilde y_{q_{\mathrm{low}}}\bigr),\\
\hat y_{q_{\mathrm{high}}} &= \hat y_{0.5} + \mathrm{softplus}\bigl(\tilde y_{q_{\mathrm{high}}}\bigr),
\end{aligned}
\end{equation}
which guarantees
\(\hat y_{q_{\mathrm{low}}} \le \hat y_{0.5} \le \hat y_{q_{\mathrm{high}}}\).  After offsetting, we apply clipping to enforce any known domain limits.

This offset‐based scheme is a simple, efficient remedy to quantile crossing without resorting to constrained optimization.  Each head $h_q$ is optimized using the standard pinball (quantile) loss. For multichannel targets $y\in\mathbb{R}^C$, we simply linearize the loss over channels. This decoupled approach is easy to implement and scales linearly in $C$, but could be extended to more sophisticated directional quantile methods \cite{kong2012quantile} in future work. Note that each quantile head is optimized independently from each other via a random task switching protocol where during training for each new batch a random choice is made to optimize the lower, median or upper quantile. 

\subsection{Performance Evaluation on Synthetic Data: Insights into System Behavior}

Here we highlight the performance of our ensemble-based model using synthetic data in a controlled setting, providing insights into the underlying system's behavior. To probe its effectiveness for denoising in a quantile regression framework, we simulate a smooth 2D surface with non-stationary, heteroskedastic, non-normal noise \emph{via} a protocol outlined in the appendix. To gauge the behavoir of our networks, we generate 10 random networks per hyperparameter combination, systematically varying the parameters alpha and gamma \-- with values equal to 0, 0.5, 1.0, 1.5 \-- as well as the network depth, ranging from 5 to 30 in steps of 5. Each network is independently trained on 8 $64 \times 64$ images, and for every model, we record emergent properties such as the number of parameters, the longest path, the average degree per node, and the record the final correlation coefficient with the ground truth data. These emergent properties are then binned into five performance categories (low, mid-low, mid, mid-high, and high), corresponding to quantile bins [0.0, 0.10, 0.30, 0.70, 0.90, 1.0]. The relationship between hyperparameters, emergent properties, and model performance is visualized using a parallel coordinate plot, Supplementary Figure 1. 

One important takeaway from this analysis is that there no single, clear factor that dictates the optimal architecture for these types of neural networks. Instead, the simulations indicate that a sufficiently large network is necessary to provide a reasonable approximation of the underlying signal, and networks with too few parameters or low average degrees consistently underperform. This behavior mirrors the phase transition phenomenon observed in random graph theory, particularly the sudden emergence of a giant component. Similarly, the neural networks here exhibit a comparable pattern: below a certain complexity threshold, performance is poor, but as complexity increases past a specific point, performance improves dramatically, until it saturates. 

This suggests that neural network functionality doesn’t improve gradually with increasing complexity but undergoes a relatively sudden transition from non-functional to functional once a critical threshold is reached, a computational phase transition that dramatically shifts the network’s capabilities. Viewing network performance through this phase transition lens provides valuable insights for hyperparameter tuning. Instead of exhaustively exploring the entire parameter space, the focus should be on identifying the critical threshold where networks move from sub-functional to functional. Beyond this point, further complexity yields diminishing returns and may even lead to overfitting.

\begin{figure}[!h]
\begin{center}
\includegraphics[width=0.95\textwidth]{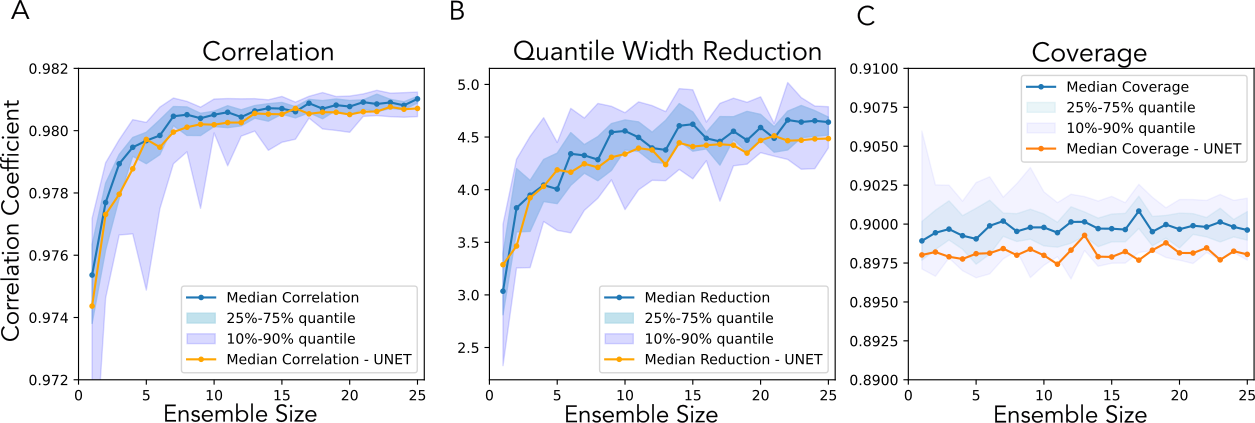}
\end{center}
\caption{Panels (A), (B), and (C) display the effect of ensemble size on model performance, focusing on three key metrics: correlation coefficient, quantile width reduction, and coverage. Panel (A) shows the median correlation coefficient as a function of ensemble size, with shaded regions representing the 25\%-75\% and 10\%-90\% quantiles. Panel (B) illustrates the reduction in quantile width, calibrated via conformal prediction, with similar quantile ranges shown. Panel (C) presents the coverage statistics for the 90\% quantile, again as a function of ensemble size. In each panel, the results for the SMSNet ensemble (blue) are compared with those for the UNET ensemble (orange), demonstrating the improved performance of the SMSNet ensemble across all ensemble sizes. The data in all panels highlight the benefits of using larger ensembles for improved accuracy and robustness in model predictions. }
\label{fig:2}
\end{figure}

In an ensemble setting, we observe an additional performance boost when combining networks. This is shown in Figure \ref{fig:2}, where we show the impact of ensemble size on the performance of the denoising. We fix the depth of the networks at 25 and measure performance by varying the ensemble size. We report the median correlation coefficient (CC) between predicted median image and the ground truth as a function of ensemble size, along with the 5\%, 25\%, 75\%, and 95\% quantiles of observed CC values over 20 independent random ensemble constructions with reported size. To evaluate the quality of our conformalized quantile estimates, we measure how much the predicted quantile intervals improve upon baseline uncertainty estimates. Specifically, we report the ratio between the 5\%-95\% quantile width of the error-free data and the mean predicted quantile width from our conformalized ensemble method. This ratio provides intuitive insight into our model's performance: a value approaching 1.0 would indicate an uninformative model yielding the widest reasonable intervals, while larger values demonstrate meaningful uncertainty quantification. We also track coverage statistics to ensure our calibrated 90\% prediction intervals maintain their intended reliability across all test cases.

The results consistently show that using the ensembles improves the mean CC by a few percentage points, and reduces quantile width by a factor of two in the most optimistic case, demonstrating the power of ensemble averaging. In order to gauge the effect of choice of convolutional neural network architecture, we apply the same analysis but replace the SMSNet with a standard UNet in the outlined framework, and rely on introducing variability via random weight initialization. Although the UNETs show signs of overfitting during training, the ensemble approach, followed by conformal calibration, helps mitigate these overfitting effects. As can be seen in Figure 1, the proposed architecture outperforms the UNet ensemble across all ensemble sizes as measured by the median criteria for both the denoised image, quantile widths reductions and coverage estimates.

\section{Results}

In this section, we demonstrate the utility of our proposed denoising framework using real scientific data, applying it to SEM-EDX and XCT imaging from a soil core obtained at the National Ecological Observatory Network (NEON) Toolik Field Station, previously used to study soil pore networks \citep{Rooney2022}. The primary goal of this study is to show how the denoising scheme, using rapid data acquisition followed by quantile regression and conformal prediction, can lead to actionable insights for more efficient downstream processes. By reducing the time and data required for high-quality analysis, we open up new possibilities for scientific exploration that were previously limited by resource constraints.

\subsection{Denoising SEM-EDX Data}

We applied our denoising framework to SEM-EDX data collected from a soil core by acquiring ten rapid 7-second exposure images, capturing elemental abundance information for $Si$, $Al$, and $Fe$ at a pixel resolution of approximately $0.5\ \mu m \times 0.5\ \mu m$. The resulting dataset, shaped as (3, 753, 1024), was split into training, testing, and calibration subsets. Since each image captured the same field of view, a high-exposure (7200-second) image was collected to serve as ground truth for evaluating denoising performance.

\begin{figure}
\begin{center}
\includegraphics[width=0.990\textwidth]{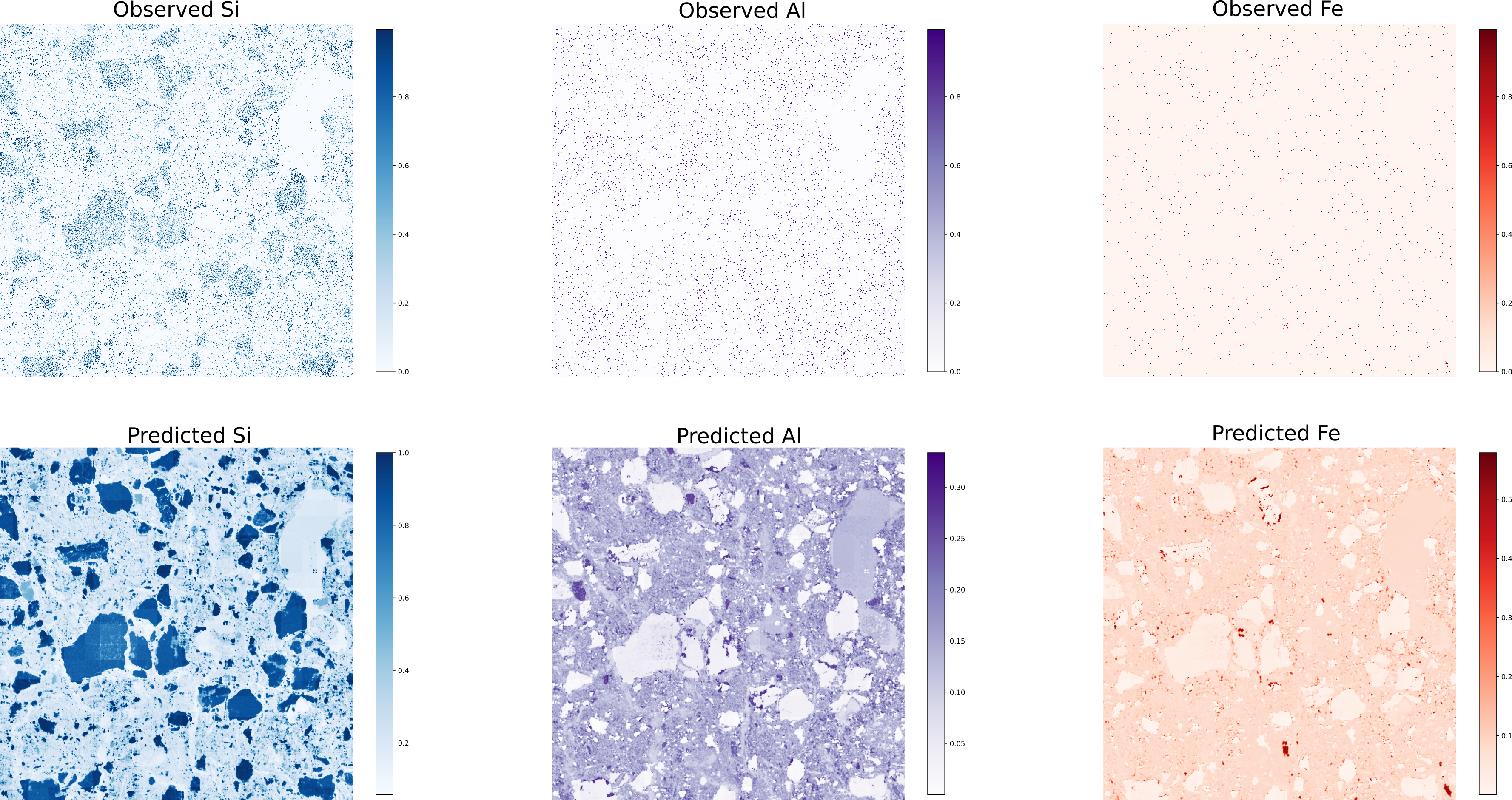}
\caption{Denoising results for SEM-EDX elemental maps of a soil core. Top: Low-exposure (7 s) input images for Si, Al, and Fe. Bottom: Predicted median outputs. Denoising reveals fine structure and material boundaries otherwise lost in noise. The displayed patch is 2.5 x 2.5 mm, an area over 50 times larger then the training data.}
\end{center}
\label{fig:3}
\end{figure}

We trained an ensemble of nine SMSNet-based convolutional neural networks, each with a depth of 15 and hyperparameters $\alpha=0$ and $\gamma=0$ and between 30k and 65k paramete. Each network produced 8 output channels, with dedicated quantile projection heads for the 25\%, 50\%, and 75\% quantiles. This quantile configuration was chosen to accommodate the extreme sparsity of the experimental signals in a low exposure setting, while still yielding actionable insights. The networks were trained for 50 epochs using the Adam optimizer with a learning rate of $10^{-3}$, yielding individual correlations of the predicted median to the 7200 second data around 89\% for the training data and 81\% for the test data.

Although the networks were trained on a relatively small area (500 $\mu m \times$ 200 $\mu m$), we deployed the ensemble to denoise a much larger region (1.5 cm$^2$) by tiling the input into 770 patches of 7-second exposures. Collecting this entire dataset at the high-quality 7200-second exposure level would have required over two months. In contrast, our method delivers actionable denoised outputs within minutes, complete with well-calibrated uncertainty bounds. As shown in Figure \ref{fig:3}, the predicted median images of an interior section of the sample reveal a meaningfull geophysical structure, while the quantile predictions (Figure~\ref{fig:S7}) provide insights into signal variability, localized uncertainty and region of interest definition.

\subsection{Emergent Latent Representations in XCT: From Denoising to Semantic Structure}

While the primary goal of our framework is denoising with uncertainty quantification, its application to volumetric X-ray Computed Tomography (XCT) data revealed a surprising emergent behavior: the formation of semantically meaningful latent representations. Trained only on noisy data, and without any labeled segmentation masks, the model learns to encode distinct structural patterns—enabling self-supervised segmentation of complex material features. These results demonstrate that denoising and representation learning can arise jointly in the same training process.

We applied our ensemble framework to XCT scans of a soil core. Two datasets were collected: a fast scan with 1,000 projections acquired over 1.5 hours, and a higher-quality scan with 3,000 projections acquiered over 5 hours. These scans served as our high-noise and medium-noise training pair. Full acquisition details are provided in the Appendix.

To accommodate the 3D nature of the data, we used 3D convolutions in our base SMSNet architecture. We trained an ensemble of 12 randomly structured networks (depth 15, $\alpha = 0$, $\gamma = 0.75$), each producing an 8-dimensional latent vector per voxel and predicting the 5\%, 50\%, and 95\% quantiles via independent projection heads. The resulting neural networks contained from 80k to 150k parameters and were trained for 50 epochs with a fixed learning rate of $1 \times 10^{-3}$ on an NVIDIA A100 GPU.

The full XCT volume ($128 \times 512 \times 1300$ voxels) was split into overlapping $64^3$ cubes with 32-voxel stride. We used 900 cubes for training and 900 for validation, with conformal calibration applied to a separate disjoint subset.

As shown in Supplementary Figure~\ref{fig:S6}, the denoised median reconstructions are smooth and preserve structure, while the calibrated quantile intervals capture the underlying variability. These denoised volumes enable rapid, low-dose scanning strategies to assess data quality in near real-time, offering a 3$\times$ reduction in acquisition time, while retaining actionable structural detail. 

\begin{figure}
\begin{center}
\includegraphics[width=0.95\textwidth]{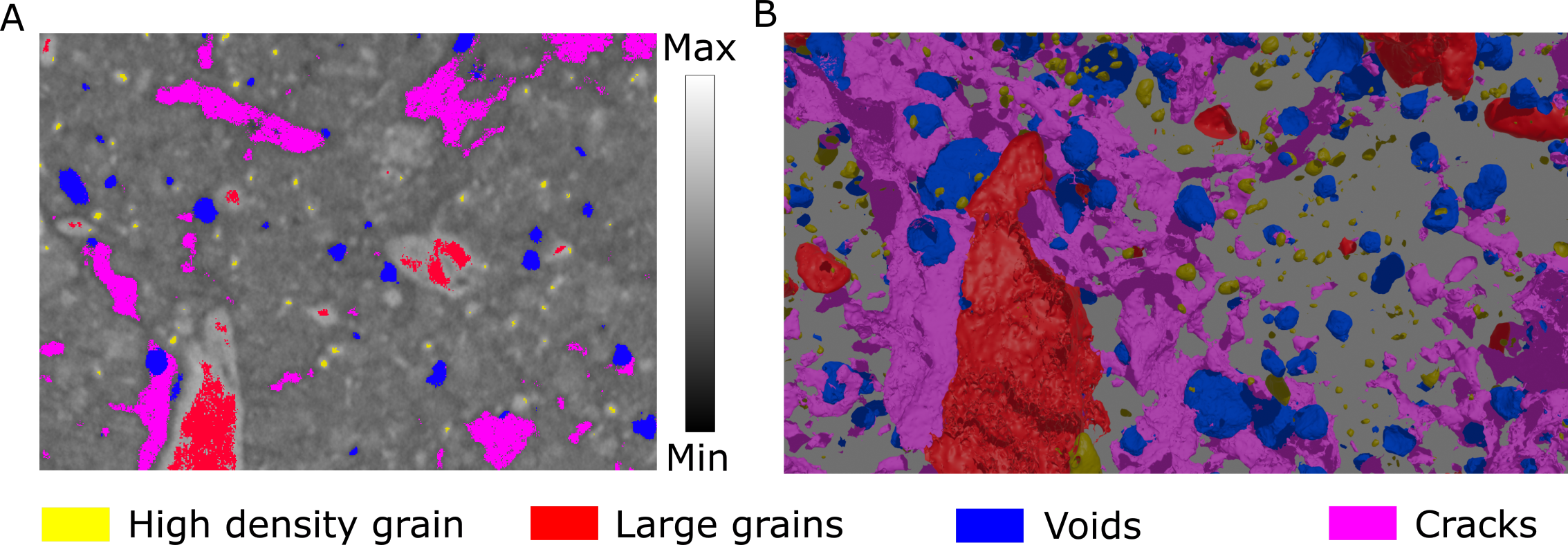}
\caption{Emergent segmentation from latent space analysis. (A) XCT slice and (B) 3D rendering with clustered latent tokens.}
\end{center}
\label{fig:4}
\end{figure}

However, the most compelling result is the structure revealed in the \emph{ensemble latent space}. After stacking all per-voxel latent vectors across trained networks, we applied singular value decomposition (SVD) followed by $k$-means clustering to the top 40 components. This unsupervised analysis uncovered spatially coherent morphological classes—including fine-grained inclusions, large grains with internal variation, voids, cracks, and transition zones—without any manual annotations or feature engineering, Figure~\ref{fig:4}. These clusters align with interpretable physical properties and can be viewed as an emergent segmentation of the sample. A further analysis highlighting that the semantic classes extend beyond density thresholding, is provided in the appendix.

These findings suggest that the latent space of even relatively small, task-specific models can encode semantically rich representations of material structure in a self-supervised manner. Such representations hold potential for downstream tasks like unsupervised segmentation, region prioritization for re-scanning, or training interpretable scientific foundation models. Similar emergent behavior was observed in the latent structure of the SEM-EDX data, as described in the Supplementary Material.

\subsection{Limitations}

Despite our promising results, several limitations warrant discussion.

\emph{Quantile-based Uncertainty Quantification:} Our conformal quantile approach guarantees coverage percentages but provides no information about the magnitude of exceedances outside prediction intervals, making it difficult to characterize extreme deviations. Future work could incorporate extreme value theory to better represent predictive distribution tails.
\emph{Conformal Prediction Assumptions:} Conformal prediction's exchangeability assumption may be violated when users selectively sample regions of interest, potentially undermining coverage guarantees. Adaptive experimental designs would require recalibration with new data to maintain valid uncertainty estimates. 
\emph{Generalizability:} The emergent representations demonstrated here are likely sample-specific, and their transferability to specimens with different compositions remains unproven. While consistent within a single dataset, a true foundation model would require training across diverse specimens and conditions. 
\emph{Computational Efficiency:} For extremely large datasets, concatenating latent representations across ensemble members becomes memory-intensive. Although we propose distributed projection techniques, further optimization is needed for scaling to gigapixel images or teravoxel volumes.
\emph{Training Data Requirements:} Our approach requires paired low/high-quality measurements from identical fields of view, which may be impractical for rare or time-sensitive specimens. Self-supervised or transfer learning approaches could expand applicability across imaging modalities. 

Despite these limitations, our approach represents a step forward in uncertainty-aware scientific imaging, offering a practical framework for making better use of limited experimental resources while providing reliable guidance for subsequent measurements and simultaneously reducing downstream annotation efforts.

\subsection{Discussion \& Conclusions}
The results of this study demonstrate that our ensemble-based denoising framework, which combines quantile regression with conformal prediction, offers a powerful and data-efficient approach to improving the quality and interpretability of scientific imaging data. In both SEM-EDX and XCT applications, we show that rapid, low-quality scans can be transformed into high-confidence estimates of the underlying structure, with calibrated uncertainty. This capability enables one to identify regions of interest in less time, decide whether further data collection is necessary, and prioritize resources for the most valuable follow-up measurements. The ability to produce calibrated uncertainty bounds around each prediction ensures that downstream usage \-- whether human analysts or machine learning models \-- is informed not just of what the model predicts, but how reliable that prediction is.

In the course of developing this denoising pipeline, we observed a surprising and valuable emergent behavior: the latent space representations learned by the ensemble encode semantically rich structure about the data itself. While our primary objective was signal recovery and uncertainty quantification, the process of training lightweight, randomly structured networks to perform quantile regression led to the formation of coherent latent vectors that consistently capture spatial and chemical features across ensemble members. By stacking these latent vectors, applying singular value decomposition, and clustering the dominant components, we uncover interpretable patterns in the XCT and SEM-EDX data that were not explicitly annotated. This structural disentanglement arises purely from the architecture and learning objective, highlighting the dual benefit of our approach: fast, uncertainty-aware denoising and the emergence of unsupervised, scientifically and semantically meaningful representations.

This observation has important implications for the future of segmentation in structural sciences. Foundational models such as the Segment Anything Model (SAM) rely on vast image–mask training sets, assuming clear and well-defined object boundaries. However, in many scientific domains \-- especially in (geo)biological imaging \-- segmentation boundaries are hierarchical, ill-defined, or application or purpose-specific. This approach hints at a way to discover internal structure of relatively small datasets without requiring hand-drawn masks, leveraging only the intrinsic consistency and redundancy present in the data itself.

In summary, we have presented a denoising and uncertainty quantification framework that is robust, data efficient, and capable of extracting rich, interpretable structure from noisy scientific data. It enables faster imaging, better resource allocation, and opens the door to self-supervised segmentation pipelines grounded in the latent structure of the data. These findings pave the way for embedding lightweight, task-adaptive representation learning directly into the acquisition and analysis loop of scientific imaging workflows.

\begin{ack}
We gratefully acknowledge the support of this work by the Laboratory Directed Research and Development Program of Lawrence Berkeley National Laboratory under US Department of Energy contract No. DE-AC02-05CH11231. Further support originated from the Center for Advanced Mathematics in Energy Research Applications funded via the Advanced Scientific Computing Research and the Basic Energy Sciences programs, which are supported by the Office of Science of the US Department of Energy (DOE) under contract No. DE-AC02-05CH11231, and from the National Institute of General Medical Sciences of the National Institutes of Health (NIH) under award 5R21GM129649-02. We acknowledge the use of resources of the National Energy Research Scientific Computing Center (NERSC), a U.S. Department of Energy Office of Science User Facility operated under Contract No. DE-AC02-05CH11231, under NERSC allocation m4055.
\end{ack}

\bibliography{bibli}
\bibliographystyle{unsrt}

\clearpage
\newpage

\appendix

\appendix  
\renewcommand{\thefigure}{S\arabic{figure}}
\setcounter{figure}{0}  

\section{Technical Appendices and Supplementary Material}
The appendices that follow provide comprehensive technical details that support and extend the core findings presented in the main text. We begin by explaining the stochastic network generation procedure that creates our diverse ensemble members, followed by the process for synthesizing controlled test data. We then provide extensive experimental details for both the synthetic benchmarks and real-world scientific imaging applications, including hyperparameter selection criteria, training protocols, and inference strategies. For each imaging modality (SEM-EDX and XCT), we include detailed information about data acquisition, preprocessing, and evaluation metrics. Additionally, we present supplementary analyses that further illustrate the emergent structural and compositional features discovered in the latent space, reinforcing our claim that meaningful representations arise naturally from the denoising objective. These technical details are provided to ensure reproducibility and to offer deeper insights into both the methodology and results.

\subsection{Random Graph Generation}
We build an SS‐DAG $G=(V,E)$ with
\begin{equation}
V = \{I,1,2,\dots,D,O\}
\end{equation}
as follows:

\begin{enumerate}
  \item \textbf{Intermediate‐to‐intermediate edges.}  For each $k=1,\dots,D$:
    \begin{itemize}
      \item Draw its out‐degree $d_k$ from the discrete distribution on $\{1,\dots,D-k\}$ with
      \[
        P(d)\;=\;\frac{e^{-\gamma\,d}}{\sum_{d'=1}^{D-k}e^{-\gamma\,d'}}.
      \]
      \item Sample $d_k$ distinct targets $k'\in\{k+1,\dots,D\}$ with
      \[
        P(k')\;\propto\;e^{-\alpha\,(k'-k)},
      \]
      and add each edge $(k\to k')$ to $E$.
    \end{itemize}

  \item \textbf{Input‐to‐intermediate edges.}
    \begin{itemize}
      \item Always add $(I\to 1)$.
      \item For $k=2,\dots,D$, add $(I\to k)$ with probability $P_{IL}$.
    \end{itemize}

  \item \textbf{Intermediate‐to‐output edges.}
    \begin{itemize}
      \item Always add $(D\to O)$.
      \item For $k=1,\dots,D-1$, add $(k\to O)$ with probability $P_{LO}$.
    \end{itemize}

  \item \textbf{Optional direct skip.}  If the flag $P_{IO}$ is true, add the edge $(I\to O)$.

  \item \textbf{Isolation check.}  For each $k=1,\dots,D$, if $k$ has no incoming edges add $(I\to k)$; if $k$ has no outgoing edges add $(k\to O)$.

  \item \textbf{Assign dilations and kernel sizes.}  For every edge $(u\to v)\in E$:
    \[
      \begin{cases}
      \delta_{u\to v}=1,\;\text{kernel size}=1, & \text{if }u=I\text{ or }v=O,\\
      \delta_{u\to v}\sim\text{UserList},\;\text{kernel size}=3, & \text{otherwise.}
      \end{cases}
    \]
\end{enumerate}

Each directed edge in the SS-DAG represents a computational operator, typically a convolution with user-specified kernel size and dilation followed by an activation function. The nodes themselves act as junction points that concatenate all incoming feature maps along the channel dimension. While our experiments here use a fixed number of output channels per convolution, the object design allows either specifying channel counts deterministically to a fixed value or sampling them at random from a user-provided list.

At runtime, we first compute a topological ordering of the SS-DAG, which yields a linear sequence of nodes such that every node’s inputs have been produced by earlier steps. We then execute the computational graph in that node order by applying its incoming edge operations, concatenate the resulting feature maps, and store the result in a memory cache that can be picked up by downstream nodes when required. Three distinct sample networks are shown in Figure \ref{fig:S1}, highlighting the stochastic nature of the generative process. Note that the 2D convolutional kernels can be replaced with 3D kernels for volumetric data.

\subsection{Synthetic Test Data Generation}
We synthesize paired “ground truth” \(S\in\mathbb{R}^{64\times64}\) and “observed” \(O\in\mathbb{R}^{64\times64}\) images as follows.  
First, a zero‐mean, unit‐variance random surface \(S\) is obtained by drawing Fourier coefficients \(c_{k,\ell}\sim\mathcal{N}(0,1)+i\,\mathcal{N}(0,1)\), 
zeroing those with \(k^2+\ell^2>\mathrm{order}^2\), enforcing \(c_{-k,-\ell}=\overline{c_{k,\ell}}\), and applying an inverse FFT.  In this study, $\mathrm{order}$ was set to 4. 
We then define a nonlinear modifier:
\begin{equation}
M(x)\;=\;\frac{1}{1 + \exp\!\bigl(-\kappa\,(x-\tau)\bigr)},
\end{equation}
with shift \(\tau\) and steepness \(\kappa\), and form the warped surface 
\begin{equation}
\tilde S = S \,\times\, M(S).
\end{equation}
$\tau$ and $\kappa$ were set to 0 and 2 respectively. Finally, we corrupt \(\tilde S\) with both additive and signal‐dependent noise:
\begin{equation}
O = \tilde S \;+\;\varepsilon_1 \;+\;\lvert \tilde S\rvert\,\sqrt{\varepsilon_2^2}\,M(\tilde S), 
\quad 
\varepsilon_1,\varepsilon_2\sim\mathcal{N}(0,\sigma^2),
\end{equation}
where \(\sigma=1.0\).  
The resulting data consist of relatively smooth surfaces \-- reflecting many imaging modalities where features exceed the resolution limit \-- and has an heteroskedastic, non‐stationary, asymmetric noise model.  

We train our model on 8 64 x 64 pixel image pairs, and 8 image pairs for conformal calibration purposes.

\subsection{Synthetic Data Benchmark \& Hyperparameter Sweep}
\begin{figure}[H]
\begin{center}
\includegraphics[width=0.995\textwidth]{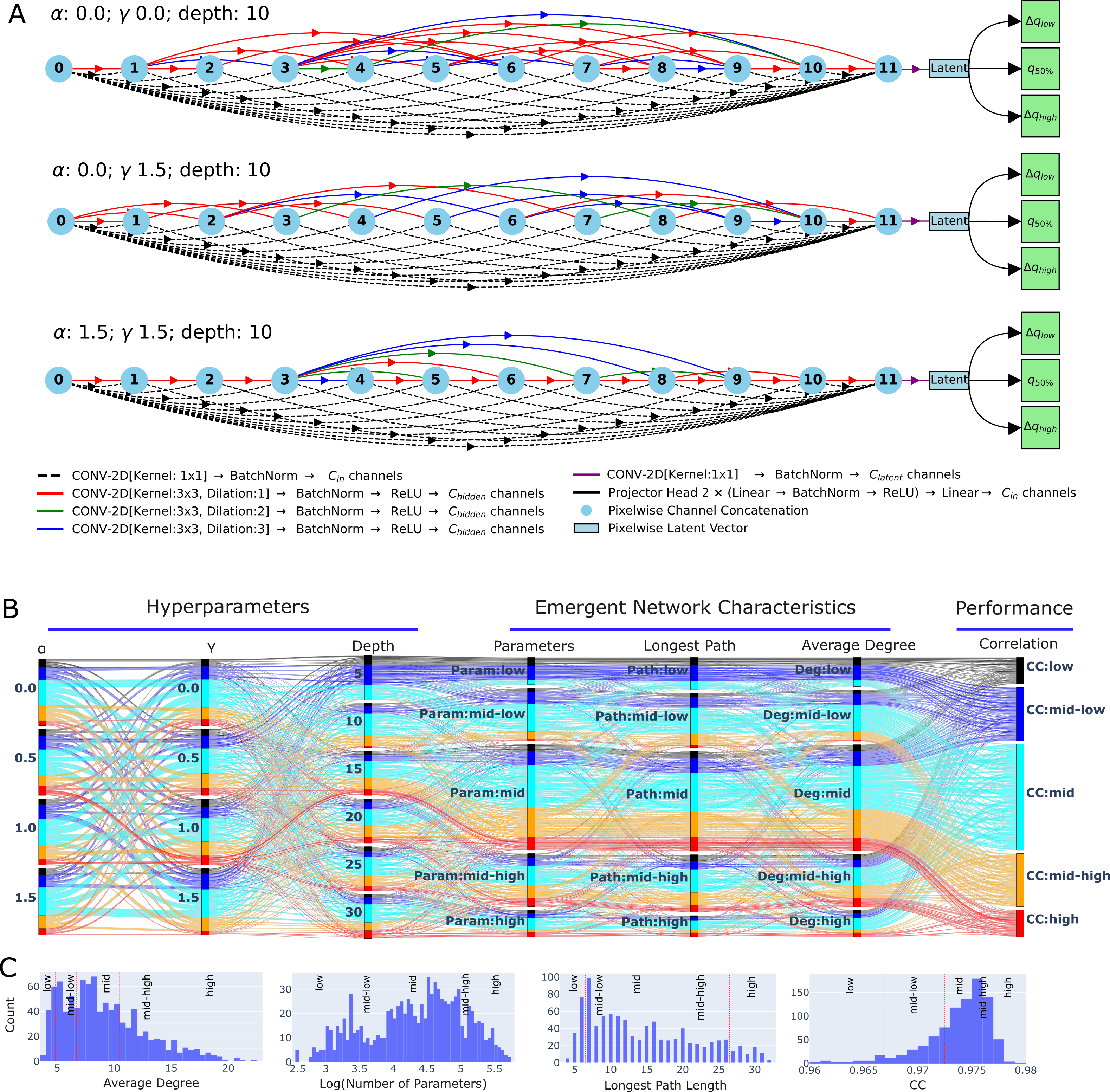}
\end{center}
\caption{(A) Visual representation of network architectures under different hyperparameter settings $\alpha$ and $\gamma$. The networks consist of an input (0), output (11) and 10 intermediate nodes (1-10) with various dilated convolutions represented by colored arrows. Black dashed lines represent skip connections between the input and hidden nodes, and hidden and output nodes. As $\alpha$ increases from 0.0 to 1.5, connections become increasingly localized between neighboring nodes. Increasing $\gamma$ to 1.5 reduces the overall connection density. Each network feeds into a latent space that produce a median predictions and quantile offset values. B. Parallel coordinate plot showing relationships between hyperparameters $\alpha$, $\gamma$, depth,  emergent network characteristics (parameter count, longest path, average degree), and performance (correlation). Colors represent performance quantiles, with warmer colors (orange/red) indicating superior performance. Networks with moderate complexity achieve optimal results. C. Histograms showing the distribution of network properties across performance bands. While there is no clear optimal performance setting, low complexity networks systematically under-perform as compared to larger networks.}
\label{fig:S1}
\end{figure}

To better understand the impact of hyperparameters $\alpha$ and $\gamma$, we visualize representative networks created under different parameter settings (Fig. \ref{fig:S1}-A). When $\alpha$ is $0.0$, there is no bias toward generating nearest-neighbor connections between nodes, resulting in networks with connections spanning across distant nodes. As $\alpha$ increases to 1.5, we observe a clear preference for shorter connections, with most links forming between adjacent or nearby nodes. When $\gamma$ is increased to 1.5, this pattern is further emphasized, as we generate networks with lower overall degree. This demonstrates how these two hyperparameters work together to control both the overall topology and network density. The performance of these networks on our synthetic data is visualized in the parallel coordinate plot (Fig. \ref{fig:S1}-B), with color schemes associated with quantiles of performance as outlined in the main text. Our analysis reveals that networks that are too small fail to achieve satisfactory performance. Only when the network reaches an appropriate size do we begin to observe networks that rank in the top 10\% of performers. Figure \ref{fig:S1}-C provides deeper insights in the distribution of the underlying network characteristics and visualizes the binning of the emergent properties and performance of the network.

\begin{figure}[H]
\begin{center}
\includegraphics[width=0.995\textwidth]{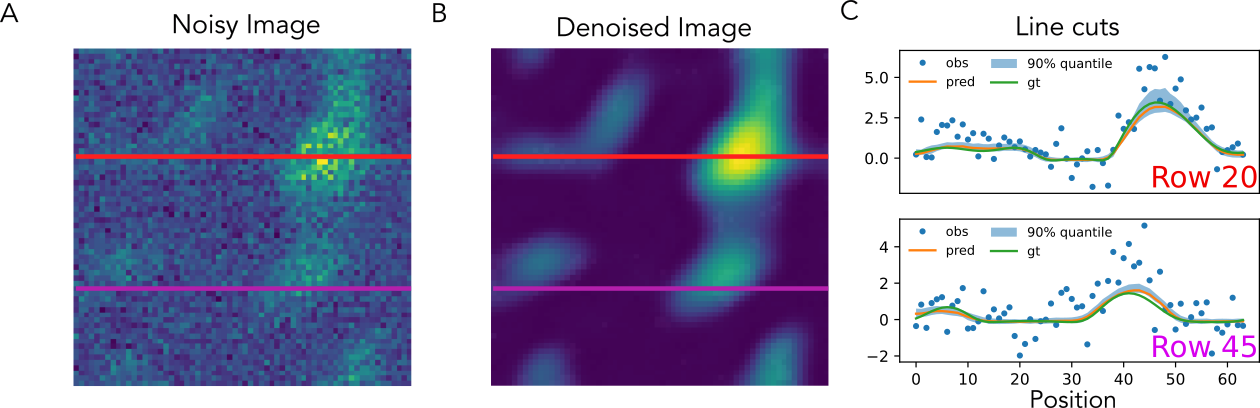}
\end{center}
\caption{Panel (A) shows a noisy image with overlaid horizontal lines indicating regions for line cut analysis. Panel (B) presents the denoised version of the same image, illustrating the improvement in image quality after applying the denoising model. In Panel (C), line cuts are shown for two specific rows (Row 20 and Row 45). For each row, the observed data (blue dots), predicted values (orange line), and ground truth (green line) are plotted, with the 90\% quantile prediction shaded in light blue. These line cuts highlight how the model effectively reduces noise and aligns closely with the true signal, providing a clear visual representation of denoising performance across different regions of the image.}
\label{fig:S2}
\end{figure}

When using the networks in an ensemble setting, a substantial performance boost is obtained. A detailed illustration of this performance boost is shown in Figure \ref{fig:S2}, where we showcases the results of applying a denoising ensemble of size 25 at depth 25 to a single image. Figure \ref{fig:S2}-A shows the noisy input surface, where significant noise is visible across the image. In \ref{fig:S2}-B, we present the denoised surface, demonstrating the improvement achieved by our ensemble model. Figure \ref{fig:S2}-C provides additional insight into the adaptive nature of the quantile regression model. It displays line cuts taken from specific rows of the image (Row 20 and Row 45), with the observed data (blue dots), the model’s predicted values (orange line), and the ground truth (green line) overlaid. The 90\% quantile prediction is shaded in light blue, highlighting how the model adapts its prediction across different regions of the surface. These line cuts emphasize the model's ability to adjust its uncertainty estimates and achieve improved alignment with the true data, underscoring the effectiveness of the quantile regression approach in handling noise.

\clearpage
\newpage
\clearpage
\newpage

\subsection{SEM-EDX Imaging}
\subsubsection{Experimental Details}
Scanning Electron Microscopy (SEM) analyses were performed with a field emission FEI Helios NanoLab 600i microscope. Prior to analysis, air-dried soil from the National Ecological Observatory Network (NEON) Toolik field station, previously used to study soil pore networks \cite{Rooney2022} was screened to select soil macroaggregates ranging from 15-25 mm in size. These macroaggregates were placed in the center of a Corning Falcon 15 mL conical centrifuge tube, where they remained trapped as the tube was filled with a slow-curing epoxy resin. 

The epoxy mixture consisted of two parts by volume of EpoThin 2 resin (Buehler Inc., Lake Bluff, Illinois) to one part by volume of EpoThin 2 hardener. By mass, 100 parts of resin were mixed with 45 parts of hardener. Sufficient epoxy resin mixture was added to each centrifuge tube to completely immerse the soil macroaggregates. The filled tubes were then degassed in a vacuum desiccator, cycling between ambient pressure and -1 atmosphere for 10 cycles, and allowed to cure overnight. To maintain the orientation of the macroaggregates during sectioning, a reference mark was made along the length of the centrifuge tube. Thin wafers,  hereafter referred to as thin sections, approximately 0.2-0.25 mm thick, were cut from the epoxy-embedded macroaggregates using an Isomet 1000 diamond blade thin sectioning saw (Buehler Inc.) equipped with a 3" × 0.006" ultrathin blade (Electron Microscopy Sciences, Hatfield, PA). Deionized water was used as the cutting fluid.
Thin sections were mounted on SEM Al-stub holders using double-sided conductive carbon tape. To improve surface conductivity, the thin sections were coated with 10-20 nm of carbon using a Ted Pella, Inc. Model 108C thermal evaporator and were placed into the SEM stage adding Cu-conductive tape around them, for further increasing the conductivity. Thin sections of macroaggregate were then analyzed at a working distance of 4 mm, using a 20 kV accelerating voltage and a 2 nA probe current. Chemical imaging and spatial elemental mapping of thin sections of macroaggregates were performed using the SEM equipped with the Energy Dispersive X-Ray Spectroscopy (EDX) X-Max 80 mm$^2$ solid-state detector (SSD) from Oxford Instruments. Two different workflows were applied to collect the chemical images.
In the first workflow, data were acquired from a single 414 x 510 $\mu m^2$ region of the macroaggregate at a resolution of 832 x 1024 pixels and a pixel size of 0.25 $\mu m^2$. Two acquisition times were selected: 7 seconds (dwell time 8 µs), and 7200 seconds (dwell time 8500 $\mu s$). The 7-second acquisition was repeated 10 times, and a single 7200-second acquisition was obtained. The chemical image maps for each dataset reveals the distribution of selected soil elements, i.e. Si, Al, and Fe. The second workflow involved imaging the entire cross-section of the soil macroaggregate using the Large Area Map (LAM) application. An initial region, corresponding to the 7 second settings described above, was designated as Tile 1. The LAM application then automated the collection of 770 such 7-second tiles to cover the entire cross-section of the macroaggregate. These tiles were then compiled into a montage, showing the spatial distribution of the same soil elements (Si, Al, and Fe) across soil macroaggregate.

\subsubsection{Training \& Inference}
Training data for denoising was generated by taking the 1024 x 832 pixel image and trimming edges to yield an image of shape $(3, 1024, 753)$. As this tile was measured 10 times for seven seconds each, we obtained a data set of size (10,3,1024,753) paired with a 10-fold duplicates of single 7200-second (3,1024,753) image to yield a image or the same corresponding size as the high noise data. The training data was split up into three spatially distinct parts, using the top 420 rows for training, the next 128 rows for testing, and the bottom 205 for conformal calibration. By splitting the data spatially, rather than across batches, we ensure that the calibration data remains as independent as possible from the training data. The $(10,3,420,1024)$ training image stack was passed through \texttt{qlty} \citep{qlty2024} to obtain patches of of size $(3,128,128)$, with an overlap of 64 pixels in both spatial directions. A similar strategy was followed for the test dataset. The data for conformal calibration was partitioned without introducing overlapping windows to avoid the introduction of bias in the calibration set.  
\begin{figure}[H]
\begin{center}
\includegraphics[width=0.95\textwidth]{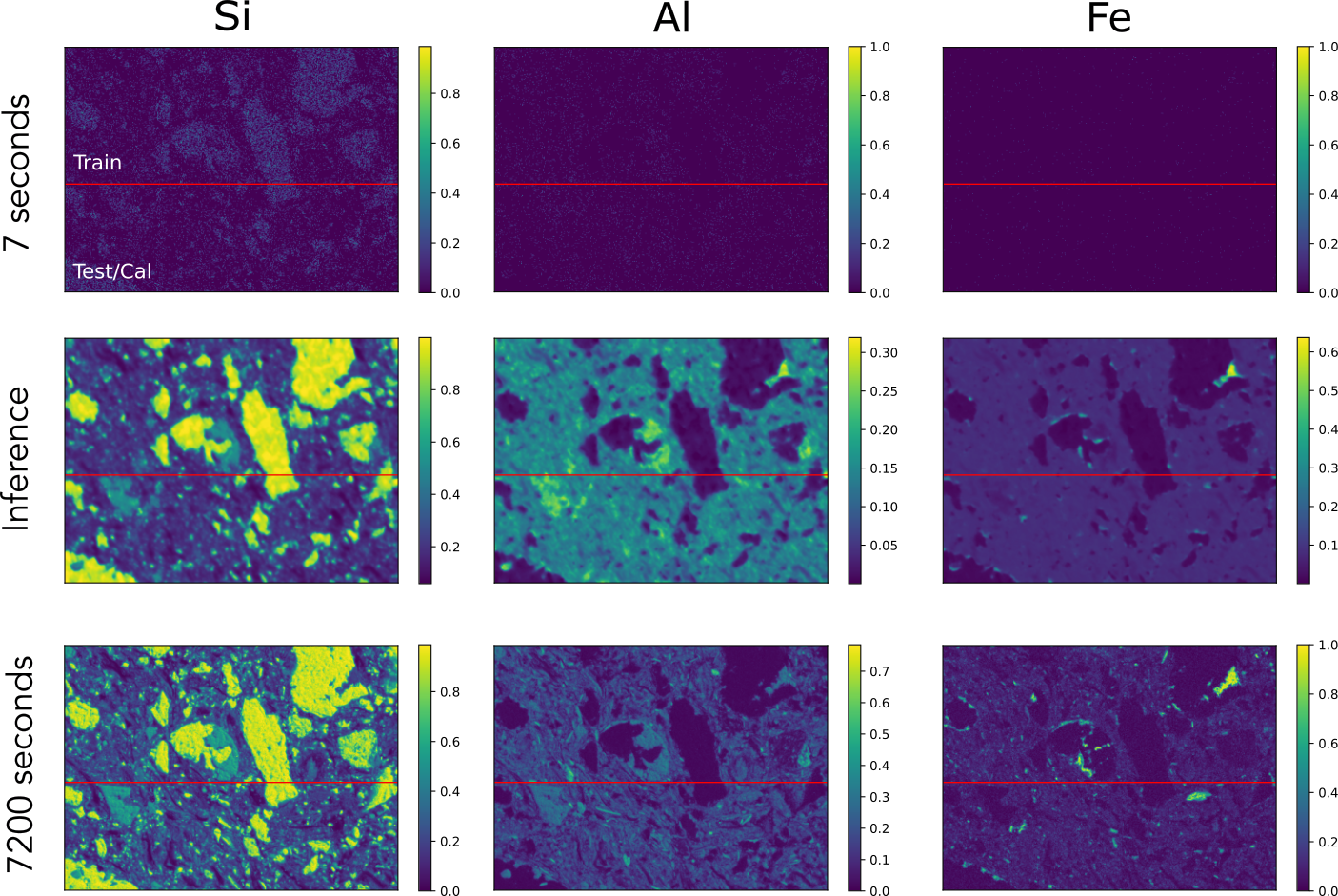}
\end{center}
\caption{Comparative analysis of elemental distributions for silicon (Si), aluminum (Al), and iron (Fe) across the different SEM-EDX channels. Top row: Low-count data obtained during short 7-second exposures, demonstrating sparse signal particularly in the Fe channel. Middle row: Inference results showing the posterior median of the reconstructed elemental distributions. Bottom row: Ground truth reference data acquired with 7200-second exposures (2 hours), confirming the accuracy of the inference procedure. Each panel displays a horizontal red line demarcating the training region (upper portion) from the test/calibration region (lower portion). Color scales indicate normalized elemental concentration values. Note that quantile limits of the posterior distributions are not visualized in this representation.}
\label{fig:S3}
\end{figure}

We trained nine independent networks of depth 15, with $\alpha$ and $\gamma$ both set to 0, dilations randomly chosen between 1 and 5, inclusive, setting the output channels for the convolutional operators between nodes to 6 and clamping degree per node between 3 and 8. The dimension of the latent space produced by the encoder was set to 8, and stepped down to 3 for the quantile projection heads as indicated in the main text. The depth of the network was chosen by limited trial and error via a number of pilot runs, yielding 15 as an acceptable depth. The number of hidden channels was expanded to 6, as past experience \-- but not grounded in firm ablation studies \-- indicate that when working with multichannel data, expanding the number of channels for the hidden layers can be beneficial. Clamping the maximum degree per node between 3 and 8 allows us to still ensure variability between networks, but have some more control over the final network size, making sure that it still runs on limited hardware. The networks generated in this manner contain between 30k and 65k parameters. The training time per image, 50 epochs using the Adam minimizer with a learning rate of $10^{-3}$ too on average 290 seconds on a RTX3090 NVIDA GPU. The average correlation coefficient between the 7200 second multichannel image and the predicted median was 89\% for the training data and 82\% for the testing data for each network.  

The results of the training are shown in Figure \ref{fig:S3}, where we compare the seven second exposure, the 7200 second exposure and the results from inference. While the resulting predicted median lacks the resolution the 7200 second image contains, it is substantially more clear than the 7 second image. Even more striking is the similarity in quality between training and testing / calibration patch, indicating that part of the overfitting is taking is mitigated by ensemble averaging.

We ran similar training rounds with a UNet of depth 3, projecting into 8 latent vectors, followed by three independent quantile heads. These network containing over 1.86 million parameters. The low amount of training data clearly resulted significant overfitting of the data, as the correlation to the ground truth of the training data was 96\%, while the correlation to the test data was 79\%. No additional regularization efforts were undertaken improve performance.  
 
\begin{figure}[h]
\begin{center}
\includegraphics[width=0.85\textwidth]{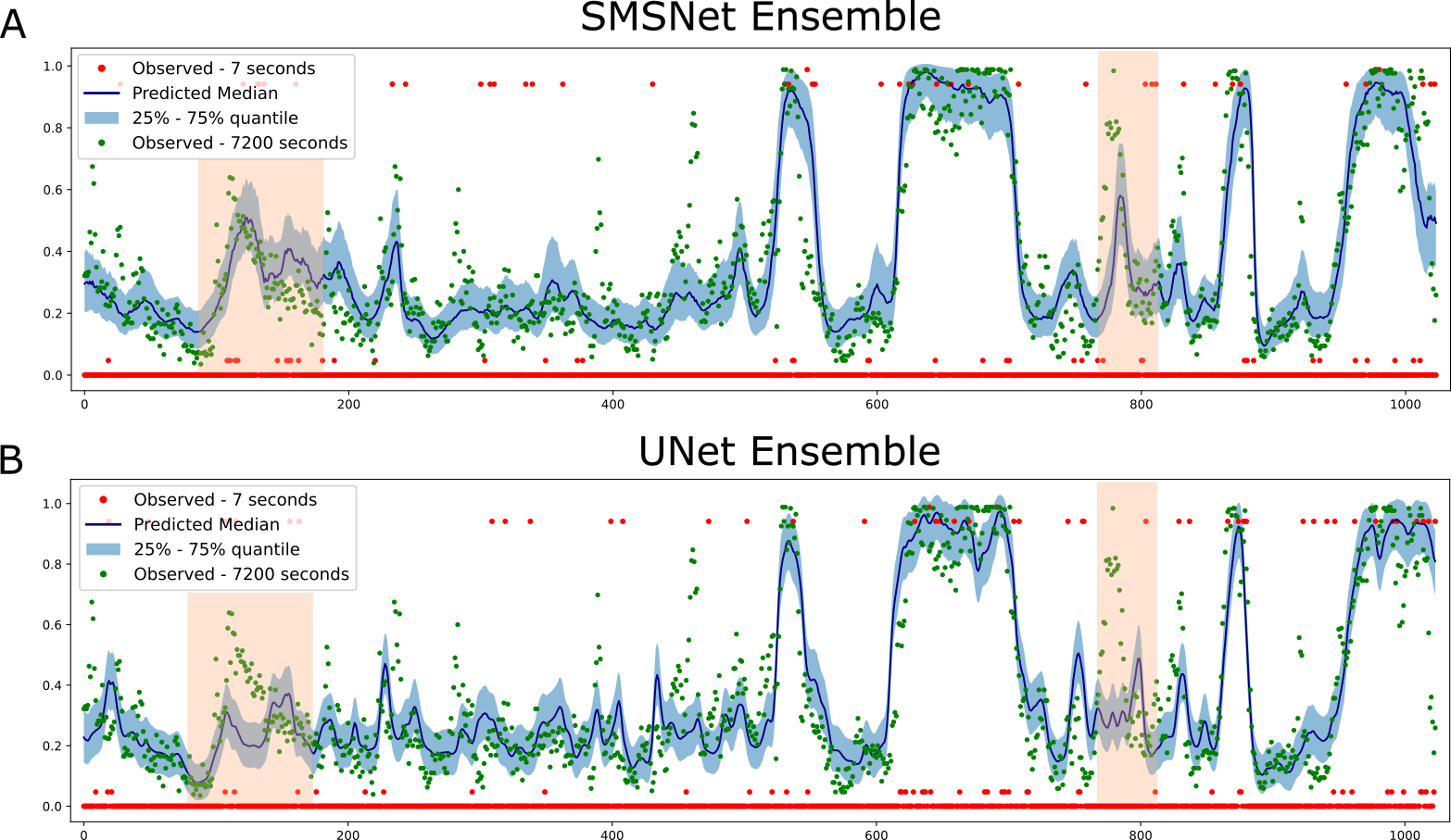}
\end{center}
\caption{Comparison of ensemble predictions for the aluminum (Al) channel from SEM-EDX data. Each panel shows observed low-exposure measurements (7 seconds, red dots), high-exposure reference data (7200 seconds, green dots), and the predicted median with the 25–75\% quantile interval (blue line and shaded region).
(A) Results from the SMSNet ensemble, which achieves a correlation coefficient of 83\% between the predicted median and the high-exposure ground truth. (B) Results from a UNet ensemble trained on the same data, with a slightly lower correlation of 79\%. Orange-shaded regions highlight areas where the two models diverge significantly. Despite similar quantile coverage, the UNet ensemble exhibits signs of overfitting, leading to qualitatively less accurate median and quantile predictions in these regions.}
\label{fig:S4}
\end{figure}

To visualize the practical consequences of overfitting in high-capacity models, we compared the performance of our lightweight SMSNet ensemble to a conventional UNet ensemble on real SEM-EDX data. As noted in the main text, UNets possess a substantially higher number of parameters, making them prone to overfitting in data-limited regimes. While conformal calibration on a disjoint dataset corrects for this overfitting in terms of overall quantile coverage, it does not resolve local miss-calibrations or erratic median predictions. Supplementary Figure \ref{fig:S4} illustrates this: although both ensembles show comparable coverage between the 25th and 75th percentiles, the UNet ensemble exhibits qualitatively noisier behavior, particularly in regions highlighted in orange where its predictions deviate substantially from the ground truth, exemplified numerically in lower correlation coefficients between the predicted median and 7200 second exposure. In contrast, the SMSNet ensemble more consistent median estimates, reinforcing the benefit of compact, diverse architectures in producing reliable uncertainty-aware predictions.

\begin{figure}[H]
\begin{center}
\includegraphics[width=0.995\textwidth]{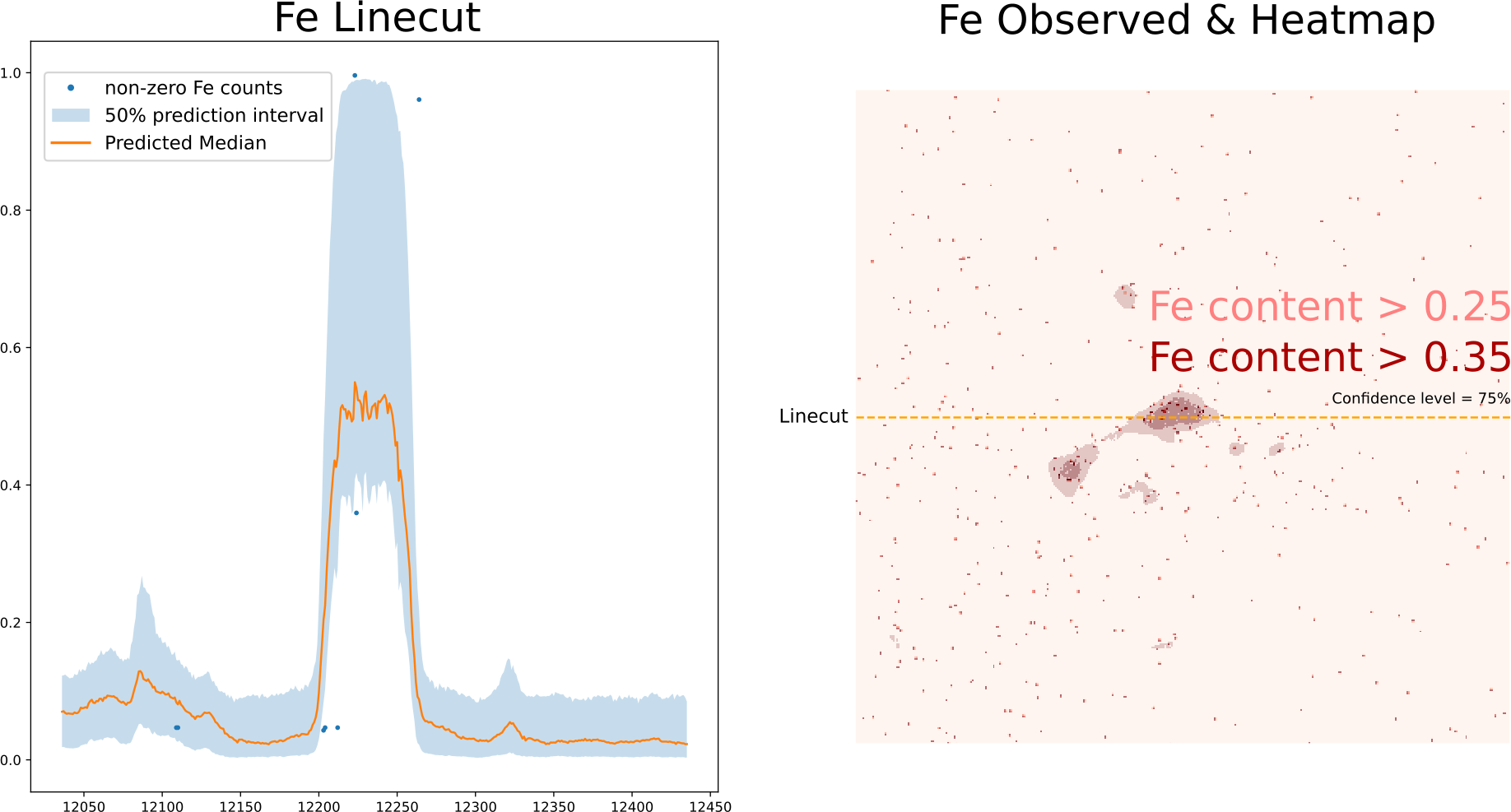}
\end{center}
\caption{(Left) Linecut through a region of the SEM-EDX data showing observed non-zero Fe counts (blue dots), the predicted median Fe signal (orange line), and the 50\% prediction interval (blue shaded region). This highlights the model’s ability to localize high-intensity Fe signals with calibrated confidence.
(Right) Confidence map indicating the probability that Fe content exceeds specific thresholds at each pixel. Darker shades correspond to higher predicted concentrations, with overlaid labels marking pixels where the Fe content is expected to exceed 0.25 or 0.35 with at least 75\% confidence. This representation enables automated targeted remeasuring of enriched regions without requiring manual inspection.}
\label{fig:S5}
\end{figure}

As demonstrated in the main text, our method yields high-quality median estimates for denoised signals. Here, we highlight how the predicted quantiles—specifically the conformalized lower bounds—can be used to guide downstream decisions. In scenarios where the goal is to identify regions with elevated Fe content, we can leverage the quantile estimates to ask whether a given threshold is exceeded with high confidence. By examining whether a specific Fe concentration lies above the 25\% lower bound, we can construct level sets that pinpoint regions of likely enrichment. This is illustrated in Figure \ref{fig:S5}, where we visualize areas predicted to contain an Fe content thresholds of 0.25 and 0.35. These confidence-based maps can drive adaptive acquisition strategies, enabling automatic re-scanning of relevant regions without human intervention—thereby enriching experimental datasets in a principled and efficient manner.

\subsection{XCT imaging}
\subsubsection{Experimental}
The aforementioned embedded soil aggregate samples were imaged in plastic conical-bottom centrifuge tubes using XCT on an XTek/Metris XTH 320/225 kV scanner (Nikon Metrology, Belmont, CA). Data were collected at 100 kV and 400 $\mu$A X-ray power. A 0.5-$mm$ thick Cu filter was used to reduce beam hardening artifacts by blocking out low-energy X-rays. The samples were rotated continuously during the scans with momentary stops to collect each projection (shuttling mode) while minimizing ring artifacts. A total of 3000 projections were collected over 360$^{\circ}$ rotation recording 4 frames per projection with 1 s exposure time per frame. Image voxel size was 10.3 $\mu m$. The images were reconstructed to obtain 3D datasets using CT Pro 3D (Metris XT 2.2, Nikon Metrology). A similar protocol was used for the high noise dataset, using 1000 projections instead.

\subsubsection{Training and Inference}
To prepare training data for XCT denoising, we used two 3D volumes of the same soil core: one acquired with 1,000 projections over 1.5 hours (high-noise), and another with 3,000 projections over 6 hours (medium-noise). Both datasets were reconstructed to a shared shape of $(1400, 1800, 2000)$ voxels. Due to minor differences in sample positioning or reconstruction parameters between the scans, the medium-noise volume required alignment to the high-noise reference. A rigid body alignment was performed using SimpleITK, applying a Nelder–Mead optimization strategy. This yielded no detectable rotational offset, but a translational shift with a total magnitude of 3.5 pixels, which was applied prior to network training. The full dataset was trimmed down to an three slabs of $(128,500,1300)$ for training, testing and calibration purposes.

We partitioned the high-noise and medium-noise volumes into overlapping cubic patches of size $(64, 64, 64)$, with 32-pixel overlap in all dimensions, yielding a training stack of 1800 patches. The volume was split assigning the first 900 cubes for training and the final 900 cubes for early stopping and monitoring purposes. Data for conformal calibration was obtained in a similar manner, but was spatially separated to prevent leakage of image structure between training and calibration steps.

We trained 12 independent networks using 3D convolutions, each with a depth of 15. The architecture generation followed the same SMSNet-based SS-DAG approach described in the main text. Hyperparameters were set as $\alpha = 0$ and $\gamma = 0.75$, with randomly sampled dilations from the range [1, 5], hidden layer output channels set to 5. The encoder produced an 8-dimensional latent vector per voxel, followed by quantile projection heads estimating the 5\%, 50\%, and 95\% quantiles. These heads were parameterized using three-layer MLPs, with the lower and upper quantiles modeled as positive offsets from the predicted median.

Networks were trained for 50 epochs using the Adam optimizer with a learning rate of $10^{-3}$. Final models had between 10k and 240k parameters, with a median of 100k, and were trained on a single NVIDIA A100 GPU, on average taking 1 hour per model. On the validation dataset, the average correlation coefficient between the predicted median and the medium-noise reference data was around 94\% per individual models.

The results of the denoising process are shown in Supplementary Figure~\ref{fig:S6}, where we compare the high-noise input, the denoised median prediction, and a ground-truth reference. A linecut across a representative section reveals strong alignment between the model's prediction and the medium-noise ground truth, with the prediction interval offering spatially resolved confidence bounds. This illustrates the ability of the network to recover structure from noisy XCT data, while preserving meaningful uncertainty estimates.

\begin{figure}[H]
\begin{center}
\includegraphics[width=0.995\textwidth]{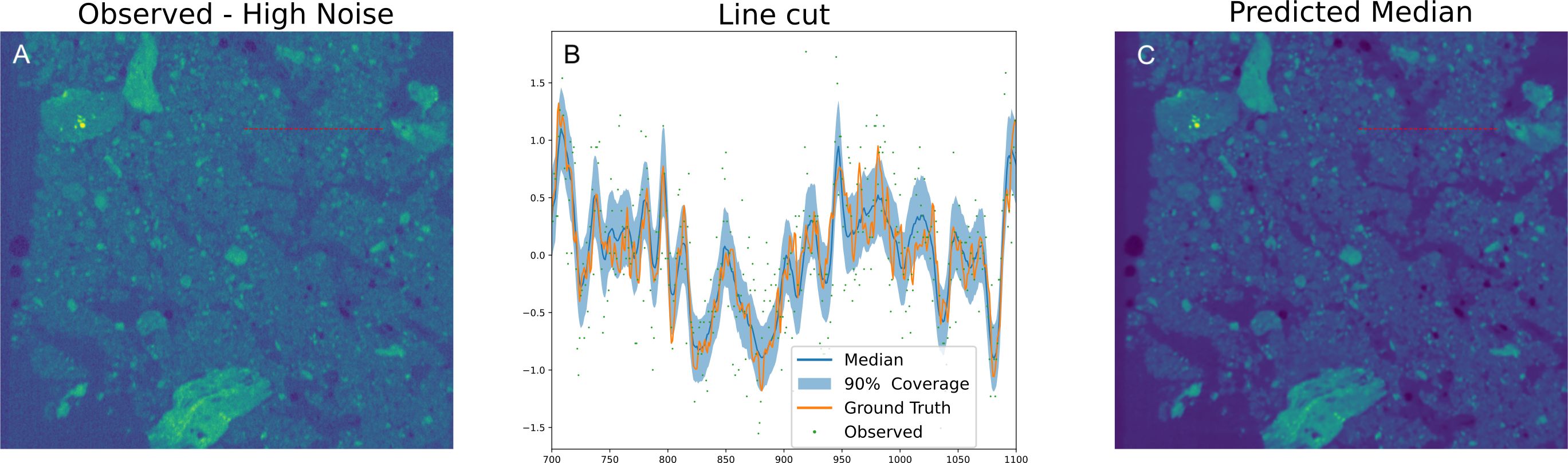}
\end{center}
\caption{A: Input XCT tomogram slice acquired under high-noise (rapid scan) conditions. B: Linecut along the red line shown in the left and right panels, comparing raw high-noise observations (green dots), reference low/medium-noise measurements (orange line), and the model's predicted median (blue line) with associated 90\% prediction interval (blue shaded area). C: Predicted median map for the same slice, denoised using our SMSNet ensemble. The comparison highlights the model's ability to accurately reconstruct signal features from high-noise input while providing calibrated uncertainty bounds across the spatial domain.}
\label{fig:S6}
\end{figure}

\subsection{Emergent Properties}
Beyond using the encoded latent vectors per pixel to obtain quantile estimates in denoising, these vectors provide a compact representation of the spatio-chemical information contained in the data. By quantizing the concatenated latent vectors via a simple k-means algorithm with 20 clusters, we can visualize the spatial distribution of these learned representations as shown in Figure S4-D. This tokenized latent vector map reveals a rich diversity of features captured by our model. Interestingly, these tokens naturally separate into those capturing morphological properties (Figure E), such as the edges of silicon-rich regions, and those representing specific chemical compositions (Figure F). This emergent disentanglement occurs without explicit supervision, suggesting that the model has learned to organize information hierarchically—distinguishing between structural boundaries and compositionally unique regions. Such representation learning not only improves denoising performance but also offers interpretable insights into material structure-property relationships that could facilitate automated characterization in future applications.

\subsubsection{Representation Learning in SEM-EDX Data}

Beyond producing quantile estimates for denoising, the per-pixel latent vectors generated by each SMSNet encode a compact and information-rich representation of the spatio-chemical structure present in the SEM-EDX data. To investigate this, we performed vector quantization via $k$-means clustering on the ensemble’s concatenated latent vectors. Each of the nine networks produced an $(8, 1024, 753)$ latent tensor, which were stacked to form a $(72, 1024, 753)$ tensor. We applied dimensionality reduction using SVD, projecting the data onto its top 20 components and clustering the resulting vectors using $k$-means with $k=20$.

\begin{figure}[h]
\begin{center}
\includegraphics[width=0.995\textwidth]{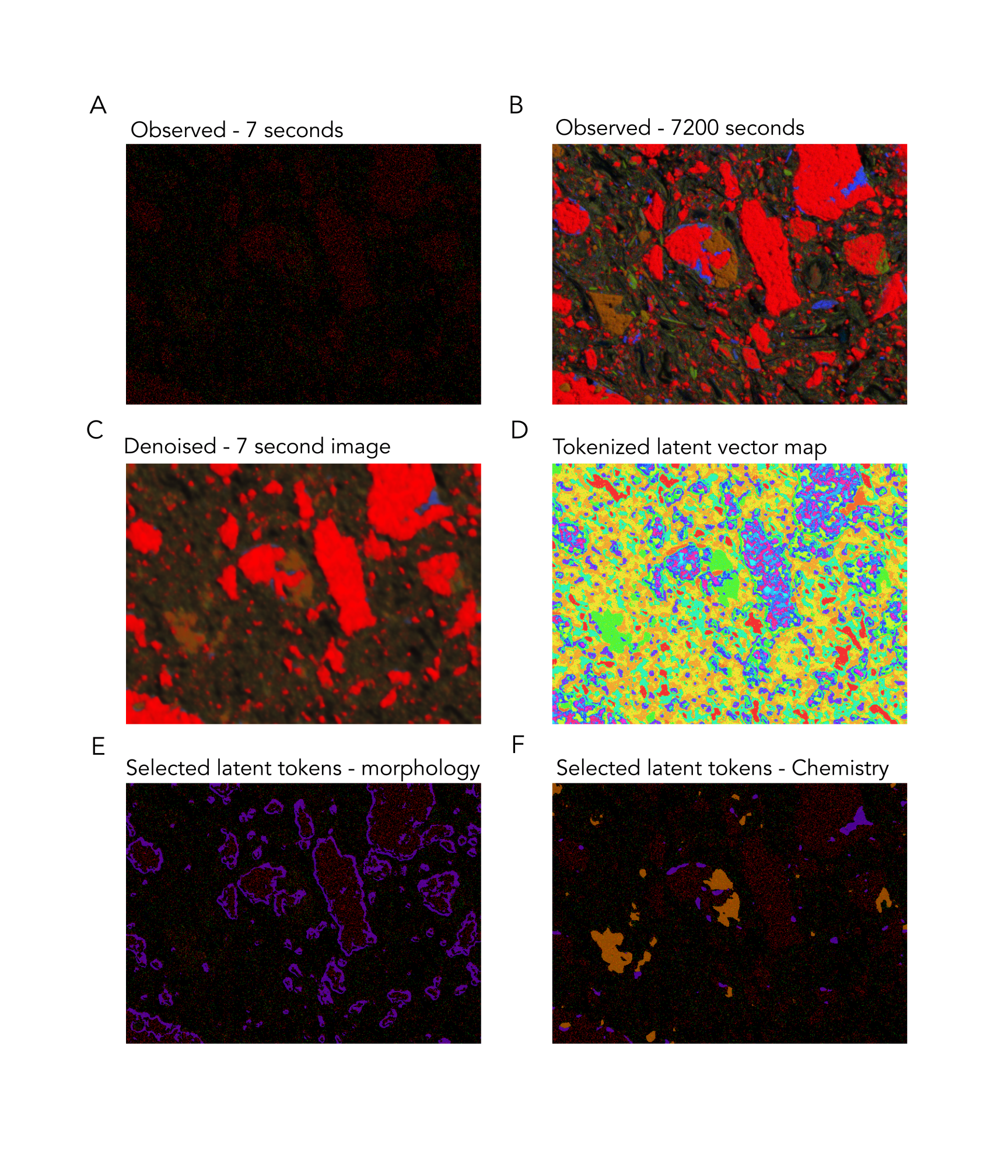}
\end{center}
\caption{Analysis beyond Quantile Regression (A) Composite RGB image from 7-second exposure showing minimal discernible features due to low signal-to-noise ratio. (B) Reference composite from 7200-second exposure revealing true material structure. (C) Denoised reconstruction from 7-second data demonstrating successful recovery of material features comparable to long-exposure ground truth. (D) Tokenized latent vector map visualizing the 20 principal components after dimensionality reduction and k-means clustering, with colors representing different latent tokens. (E) Selected latent tokens highlighting morphological features, particularly edges of silicon-rich regions (purple outlines). (F) Selected latent tokens emphasizing regions with distinct chemical composition (orange and purple regions), illustrating how the model separates structural and compositional information.}
\label{fig:S7}
\end{figure}

As visualized in Figure~\ref{fig:S7}-D, this process yields a spatial token map that highlights structurally and chemically distinct regions. Some tokens trace morphological features such as edges of silicon-rich areas, Figure~\ref{fig:S7}-E, while others highlight chemically distinct zones, Figure~\ref{fig:S7}-F. These results show that the latent space encodes disentangled features without supervision, suggesting that the network learns to hierarchically organize structure and composition. Such tokenized maps offer rich interpretability and can be leveraged for downstream tasks including segmentation or adaptive measurement.

\subsubsection*{Representation Learning in XCT Data}

Because of the vastly larger data volume in the XCT setting, we adopted an out-of-core approach for latent analysis. We concatenated the 12 ensemble latent tensors, each of shape $(8, 128, 500, 1300)$, yielding a $(96, 128, 500, 1300)$ latent volume. This dataset—containing over 83 million vectors—was stored in Zarr format and processed using Dask for scalable computation. We performed a randomized SVD on a single CPU node on Perlmutter, completing the top-20 decomposition in 6.5 minutes. The resulting $(20, 128, 500, 1300)$ projected tensor was clustered using $k$-means, producing token maps comparable in quality and semantic coherence to the SEM-EDX case, Figure~\ref{fig:4}.

\begin{figure}[h]
\begin{center}
\includegraphics[width=0.995\textwidth]{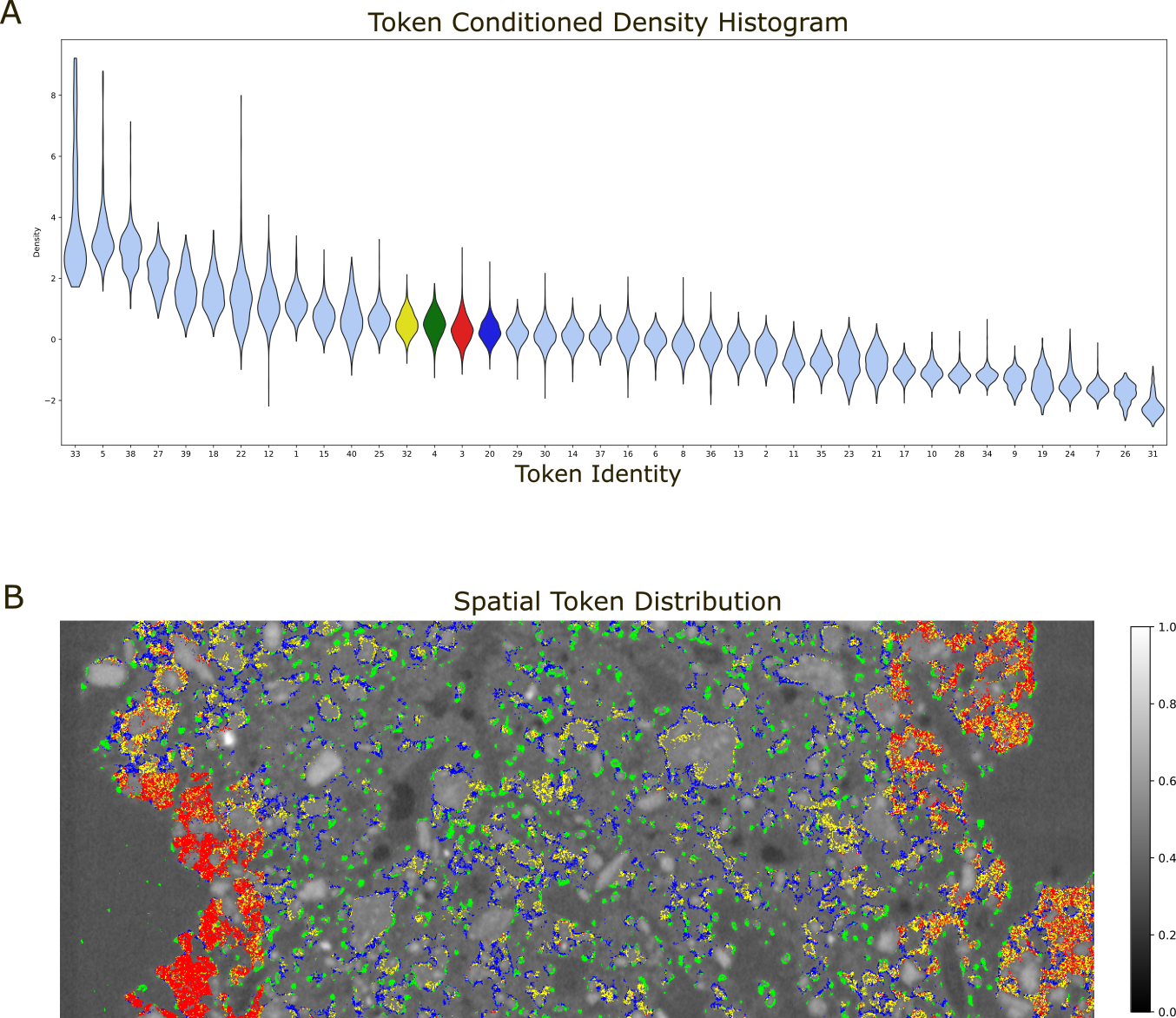}
\end{center}
\caption{(A) Token-conditioned density histogram for XCT data. Each violin plot shows the distribution of voxel intensities for one of the 40 latent tokens derived from SVD-projected latent vectors followed by $k$-means clustering. Tokens are sorted by decreasing mean density. Highlighted tokens (yellow, green, red, blue) were selected for spatial visualization in panel B. The diversity and separation of these distributions demonstrate that the tokenization reflects physically meaningful differences in material beyond mere mean density. (B) Spatial distribution of selected latent tokens overlaid on the XCT tomogram. The background shows a grayscale slice of the XCT volume, while colors indicate the spatial footprint of the four highlighted tokens from panel A. The map confirms that tokens correspond to semantically distinct features, reinforcing the interpretation that the learned latent space organizes semantic material features in an unsupervised fashion.}
\label{fig:S8}
\end{figure}

To investigate physical semantics of these tokens, we generated token-conditioned histograms of XCT voxel intensities, Figure \ref{fig:S8}. This analysis reveals that for tokens with significantly overlapping density histograms, distinct spatial patterning occurs, consistent with the interpretation that the latent space clusters material regions according to underlying structural and morphological traits.

\subsubsection{Distributed projection into shared latent subspace}

While the previous approach stacks all ensemble latent vectors into a single matrix before applying SVD and clustering, a follow-up strategy enables per-network latent compression. After computing the full SVD basis once, the right-singular vectors can be partitioned according to network-specific blocks, creating partial projectors. Each network can then independently project its latent tensor into a shared subspace. The resulting features may be averaged across models, enabling token map construction even when full stacking is computationally infeasible.

This approach also supports modular inference pipelines—individual models can be updated or swapped without recomputing the global structure. Although not employed here, it is particularly well suited for real-time or streaming settings, where each model contributes its own projected representation to a shared feature space via a simple distributed reduce operation.

Formally, consider a data matrix formed by concatenating latent representations from $N$ networks:

\begin{equation}
X_{\text{concat}} = [X_1, X_2, \ldots, X_N] \in \mathbb{R}^{d \times T}
\end{equation}

The rank-$r$ truncated singular value decomposition (SVD) of this matrix is:

\begin{equation}
X_{\text{concat}} = U \Sigma V^\top
\end{equation}

with
\begin{align}
U &\in \mathbb{R}^{d \times r}, \\
\Sigma &\in \mathbb{R}^{r \times r}, \\
V &\in \mathbb{R}^{T \times r}
\end{align}

The matrix \( V \) can be partitioned according to the original blocks of \( X_{\text{concat}} \):

\begin{equation}
V = \begin{bmatrix}
V_1 \\
V_2 \\
\vdots \\
V_N
\end{bmatrix}, \quad \text{where } V_j \in \mathbb{R}^{T_j \times r}
\end{equation}

We can express \( U \) in terms of these blocks as:

\begin{equation}
U = X_{\text{concat}} V \Sigma^{-1}
\end{equation}

Expanding \( X_{\text{concat}} \) as the sum of its components:

\begin{equation}
U = \left( \sum_{j=1}^N X_j V_j \right) \Sigma^{-1}
\end{equation}

Define the per-block projector:

\begin{equation}
P_j = V_j \Sigma^{-1} \in \mathbb{R}^{T_j \times r}
\end{equation}

Then the contribution from each network block becomes:

\begin{equation}
U_j = X_j P_j
\end{equation}

and the full left singular basis can be reconstructed as:

\begin{equation}
U = \sum_{j=1}^N U_j = \sum_{j=1}^N X_j P_j
\end{equation}

This formulation enables storing only the partial projectors \( P_j \), allowing each network to project its latent output into a shared subspace independently. The resulting design supports modular, parallel, and streaming inference pipelines.

Beyond projection and averaging, the partial projectors 
\( P_j = V_j \Sigma^{-1} \) 
naturally enable additional analysis. Since all projected outputs 
\( U_j = X_j P_j \) 
lie in a shared latent subspace, one can compute per-position variance, inter-network correlations, or attribution weights. This allows assessment of ensemble diversity and individual network contribution—e.g., by analyzing the row-wise variance or cosine similarity of 
\( U_j \) 
across models. While not explored here, this could support weighting strategies or highlight underperforming or redundant networks.

\subsection{Data \& Code Availability}
Links to data, trained networks and additonal materials are available at https://phzwart.github.io/behindthenoise/ .


\end{document}